\documentclass[10pt,logo,copyright]{nvidiatechreport}
\usepackage{float}
\usepackage{multirow}
\newcolumntype{Y}{>{\centering\arraybackslash}X}

\usepackage[numbers, sort]{natbib}
\definecolor{nvidiasapgreen}{HTML}{507D2A}
\newcommand{\bestscore}[1]{\textcolor{nvidiagreen}{\textbf{#1}}}
\newcommand{\secondscore}[1]{\textcolor{nvidiasapgreen}{\underline{#1}}}
\usepackage{amsmath,amsfonts,bm}

\usepackage{cleveref}

\makeatletter
\DeclareRobustCommand\onedot{\futurelet\@let@token\@onedot}
\def\@onedot{\ifx\@let@token.\else.\null\fi\xspace}

\def\ie{\emph{i.e}\onedot}
\makeatother

\title{Context-Matched Distillation: Teacher Causality for Autoregressive Video Distillation}

\author{
  \textbf{Hmrishav Bandyopadhyay $^{1,2}$} \quad
  \textbf{Xuanchi Ren $^{1}$} \quad
  \textbf{Zijian Huang $^{1}$} \quad
  \textbf{Jay Zhangjie Wu $^{1}$} \quad
  \textbf{Tianshi Cao $^{1}$} \quad
  \textbf{Ruilong Li $^{1}$} \quad
  \textbf{Bryan Chu $^{1}$} \quad
  \textbf{Sanja Fidler $^{1}$} \quad
  \textbf{Yi-Zhe Song $^{2}$} \quad
  \textbf{Zian Wang $^{1}$}\\
  \small $^{1}$ NVIDIA \quad $^{2}$ SketchX, CVSSP, University of Surrey\\
  {\small \url{https://hmrishavbandy.github.io/cmd-site/}}
}

\begin{document}

\maketitle

\noindent\begin{minipage}{\linewidth}
  \centering
  \vspace{-0.6cm}
  \includegraphics[width=\linewidth]{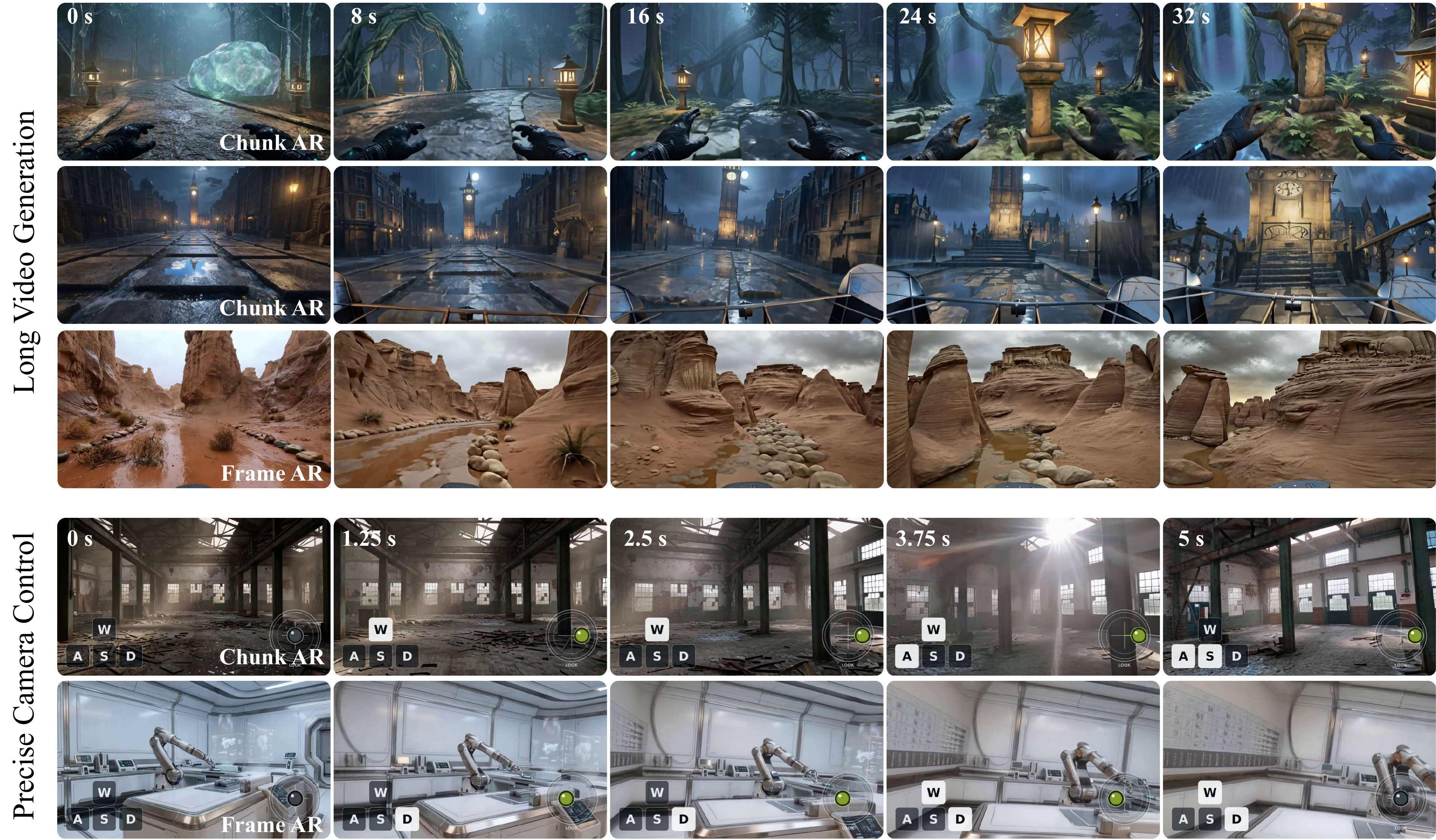}
  \captionsetup{skip=2pt}
  \captionof{figure}{\textbf{Context-Matched Distillation supports long-horizon and camera-controllable autoregressive video generation from a single image.} Starting from the conditioning image at $0$ s (left), our frame-wise (Frame AR) and multi-frame (Chunk AR) models autoregressively extend each scene. The upper rows show long-video rollouts sampled at the indicated times; the lower rows show camera-controlled rollouts guided by sequential translation (W/A/S/D) and look-direction commands. Frame AR incorporates updated controls after each latent frame, whereas Chunk AR updates controls at multi-frame block boundaries. \\ [-1.2cm]}
  \label{fig:teaser}
\end{minipage}
\vspace{0.5em}

\begin{abstract}

\noindent Interactive autoregressive video generation demands both low-latency rollouts and precise online control. Few-step distillation accelerates generation by reducing denoising steps, while online control imposes a causal constraint: frames and blocks should depend on history and controls available during generation. Existing video distribution matching distillation (DMD) pipelines, however, often supervise causal few-step students using bidirectional teachers that score complete clips. The score for a target can therefore depend on future frames and controls that were unavailable when the student generated it, misaligning teacher supervision with the student's causal information set. We introduce \textbf{Context-Matched Distillation (CMD)}, a causal DMD framework that aligns teacher supervision with the information available when each target is generated. CMD replaces bidirectional full-clip scoring with a causal teacher that evaluates each target without access to future frames or controls. The same causal teacher initializes the few-step student, establishing a consistent causal formulation across teacher training, student distillation, and inference. Beyond aligning the temporal information boundary, \textbf{Prefix Scoring} matches supervision to the student's realized rollout context by evaluating each target under the cached student-generated prefix that produced it. \textbf{Prefix Corruption} further stabilizes training by perturbing unreliable prefixes produced early in training while preserving this target--context alignment. With a simple causal formulation, CMD naturally extends to frame-wise and chunk-wise generation, long video distillation, and camera-conditioned distillation. Experiments demonstrate state-of-the-art aggregate performance among autoregressive methods on both short- and long-video benchmarks, together with substantially improved adherence to time-varying camera controls.

\end{abstract}
\abscontent

\section{Introduction}

Autoregressive (AR) video models generate visual observations sequentially, making them a natural fit for interactive generative world models \cite{yin2025causvid,huang2025selfforcing,team2026advancing,zhu2026sanawm}. To support practical interaction, these models must sustain long-horizon rollouts, accept frequent control updates, and faithfully execute time-varying control signals such as camera poses. Every generated frame or block is reused as context for the next prediction, so even small errors can compound into long-term drift in appearance, geometry, and motion.

Few-step distillation reduces the cost of each autoregressive transition by compressing multi-step denoising into a small number of student evaluations
\cite{yin2024one,yin2024improved,yin2025causvid}.
Methods such as Self-Forcing further roll out the student during training, exposing it to its own generated histories and reducing exposure bias
\cite{huang2025selfforcing,guo2025end}.
However, standard distribution matching distillation (DMD) pipelines still use a bidirectional teacher to score each completed generated clip.
A target can therefore be supervised using future frames and controls that were unavailable when the causal student produced it.
We call this the \emph{teacher--student context mismatch}.

The context mismatch is especially consequential for camera control and long video generation.
In camera-conditioned distillation, a bidirectional teacher sees the complete camera trajectory, while the student must execute each pose update online
\cite{he2024cameractrl,wang2024motionctrl}.
Future pose changes can therefore affect the supervision assigned to the current transition.
For long rollouts, treating each local clip as self-contained both discards the preceding rollout context used to generate its early targets and allows those targets to attend to later frames inside the scoring window
\cite{cai2026mode,yang2025longlive}.
In both cases, the teacher scores a target under a different context from the one in which the deployed generator acts.

To align these contexts, we introduce \textbf{Context-Matched Distillation (CMD)}, which scores each target with a causal teacher that cannot access later frames or controls.
CMD first fine-tunes a pretrained bidirectional model into a multi-step causal teacher with a Diffusion-Forcing objective over independently noised histories \cite{chen2024diffusion}, then initializes the few-step student from those weights.
During distillation, the student is rolled out autoregressively, while the frozen teacher scores its noised targets causally.
The same causal architecture and conditioning are therefore used throughout teacher training, distillation, and student inference.
As a practical side benefit, adding a new control does not require separately adapting both a bidirectional conditional teacher and a causal student, and no additional ODE-matching or consistency-distillation stage is needed for initialization
\cite{yin2025causvid,zhu2026causal,zhao2026causal++,gao2026lingbotworld2}.

Causal scoring prevents future information from affecting a target score, but it does not by itself ensure that the teacher observes the actual history that produced the target.
A naive implementation conditions on preceding noised DMD targets rather than the realized student rollout.
We therefore introduce \textbf{Prefix Scoring}, which supplies the cached student-generated prefix used to produce each target.
Early student prefixes may contain structured rollout artifacts unlike the independent corruptions seen during teacher training. \textbf{Prefix Corruption} therefore perturbs the cached prefix to reduce the teacher's sensitivity to these unreliable details while preserving causal ordering and control signals.
A block-causal attention mask computes all target-specific scores in a single teacher pass.

\begin{figure}[t!]
    \centering
    \includegraphics[width=\linewidth]{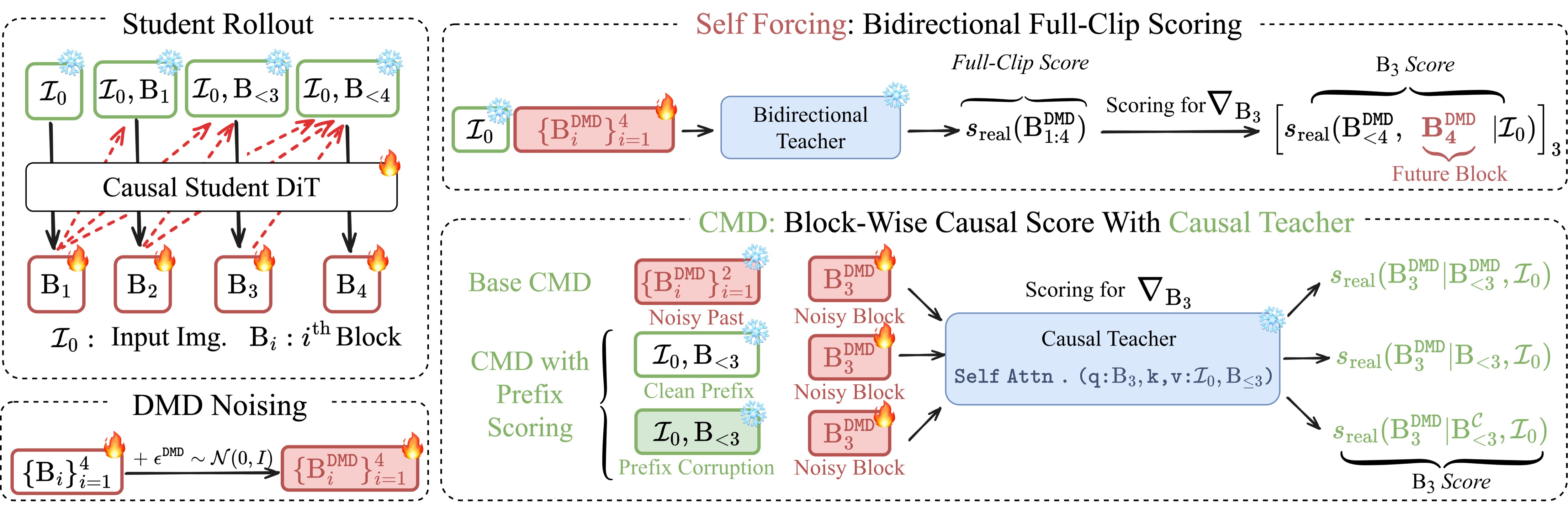}
    \caption{\textbf{Context-Matched Distillation.}
    A causal student first generates an autoregressive rollout. Self-Forcing scores its complete noised rollout with a bidirectional teacher, allowing the score of a target to depend on future blocks. CMD instead uses a causal teacher: Base CMD conditions on preceding noised DMD blocks, Prefix Scoring uses the cached clean student prefix that produced the target, and Prefix Corruption perturbs this prefix to stabilize supervision when early rollouts are unreliable.}
    \label{fig:concept}
\end{figure}

CMD's causal target--prefix scoring applies naturally to frame-wise and chunk-wise generation, bounded-context long videos, and time-varying camera control. Across short- and long-video evaluations, CMD consistently improves quality over bidirectional scoring baselines. For camera-conditioned generation, it substantially reduces camera-pose errors, confirming the importance of matching teacher supervision to the temporal information available during online generation. Ablations further isolate the contributions of causal scoring, matched rollout prefixes, and prefix corruption.

We summarize our contributions as follows:
\begin{itemize}
    \item We introduce \textbf{Context-Matched Distillation}, which uses a causal teacher to score every generated target without access to later frames or controls.
    \item We introduce \textbf{Prefix Scoring} to condition each score on the student-generated history that produced its target, together with \textbf{Prefix Corruption} for robust training on unreliable early rollouts.
    \item We apply the same formulation across frame-wise and chunk-wise generation, long-video rollout, and camera control, achieving strong short- and long-video quality and substantially lower camera-pose errors.
\end{itemize}

 \section{Related Work}
\noindent \textbf{Block-causal autoregressive video generation.}
Video diffusion pipelines traditionally use bidirectional attention across all frames \cite{wan2025wan,kong2024hunyuanvideo}, which promotes clip-level appearance and object consistency but scales quadratically with the number of tokens \cite{dao2024flashattention}.
Causal and block-causal models instead generate frames or short frame blocks sequentially, reusing cached history across generation steps \cite{gao2024ca2,yin2025causvid,huang2025selfforcing,zhou2025taming,guo2025end}. They support streaming and interactive generation \cite{team2026advancing,zhu2026sanawm,liu2026gamma}, while multi-step causal models can also be pretrained directly with flow matching or Diffusion Forcing \cite{chen2024diffusion,teng2025magi,guo2025end}. Most few-step variants are obtained by distilling a pretrained bidirectional model with a causal attention mask \cite{yin2025causvid,huang2025selfforcing,liu2025rolling,yang2025longlive,zhuang2026self} with distribution matching distillation \cite{yin2024one}, often supplemented with adversarial objectives \cite{feng2026one,lin2026autoregressive,huang2025selfforcing}. To reduce the mismatch between a bidirectional teacher and a causal student, recent works use a causal teacher to initialize the causal student through ODE matching \cite{zhu2026causal} or consistency distillation \cite{zhao2026causal++}, or to fine-tune a learned student for long-video generation \cite{chen2026contextforcing}. Unlike these approaches, we train a causal teacher for multi-step generation and use the same teacher for student initialization and distillation with causal prefixes. Our approach reduces the teacher--student context mismatch, enabling high-quality video generation with precise control.

\noindent \textbf{Long video generation.}
Long-video diffusion has been extended beyond a model's training horizon through tuning-free noise rescheduling \cite{qiu2024freenoise}, fixed-memory diagonal denoising \cite{kim2024fifo}, autoregressive short- and long-term memory modules \cite{henschel2024streamingt2v}, and Diffusion-Forcing-based infinite-length generation \cite{chen2025skyreels}.
Recent few-step autoregressive systems use rolling windows and attention sinks for real-time long rollout \cite{yang2025longlive,liu2025rolling}, while Infinity-RoPE extends the positional horizon and refreshes the cache for responsive action control at inference time \cite{yesiltepe2026infinity}.
Context-management methods compress history into a bounded token budget with importance-weighted frame packing \cite{zhang2026frame}, deep sinks and attention-based cache pruning \cite{yi2025deep}, or sink--compressed--recent cache partitions \cite{mao2026packforcing}.
Learned alternatives combine a local attention window with recurrent state-space memory \cite{yu2025videossm}, or retrieve pose- and time-indexed frames from an explicit memory bank \cite{xiao2026worldmem}.
Training-based methods further use sink tokens \cite{cui2026lol}, longer-video supervision and hierarchical denoising \cite{cai2026mode}, or a long-context teacher with slow--fast memory \cite{chen2026contextforcing}.
These methods improve the student's memory or extend the teacher's temporal reach, but distillation with a bidirectional teacher still requires scoring bounded subclips of a much longer causal rollout.
We instead query a causal teacher under exactly the bounded context used by the student, so every frame is supervised under the context that generated it without increasing the teacher's attention window.

\noindent \textbf{Camera-controlled video generation.}
Camera motion is commonly injected into video generators through spatially aligned ray maps or dedicated pose adapters, enabling trajectory-conditioned generation and separating viewpoint changes from object motion \cite{he2024cameractrl,xu2024camco,wang2024motionctrl,ren2025gen3c,shen2026lyra2}.
This line of work has been extended to iterative scene exploration over wider viewpoint ranges and more geometrically consistent novel views \cite{he2025cameractrl,xu2024camco}.
Geometry-aware alternatives transform or modulate attention using relative camera poses \cite{miyato2024gta,cape,li2026prope}, or encode positions directly in ray space \cite{yin2026raype}.
Recent causal world models extend autoregressive generation to long-horizon rollout with discrete actions and interactive 6-DoF camera control using these representations \cite{sun2025worldplay,zhu2026sanawm,team2026advancing,zhao2026minwm,gao2026lingbotworld2}.
Few-step distillation further reduces rollout latency for real-time camera-controlled novel-view generation \cite{xu2026realcam,team2026advancing,zhao2026minwm}.
However, a bidirectional teacher that receives the complete camera trajectory can use future poses when scoring the current frame, whereas an online autoregressive student only observes past and current controls.
We use a frame-relative ray map and score each camera update with a causal teacher under the generated prefix, eliminating this future-camera lookahead during distillation.
 \section{Preliminaries}
\label{sec:prelims}

\noindent \textbf{Distribution Matching Distillation.} Distribution matching distillation~\citep{yin2024one,yin2024improved} distills a few-step student generator $G_\theta$ from a multi-step teacher by matching the student distribution $p_\text{fake}$ to the teacher distribution $p_\text{real}$. DMD uses the difference between their score functions, $s_\text{fake}$ and $s_\text{real}$, as a proxy for the distribution mismatch and minimizes this. The resulting gradient is written as
\begin{equation}
    \nabla_\theta \mathcal{L}_{\mathrm{DMD}}
    =
    -\mathbb{E}_{x \sim p_\text{fake}}
    \left[
        \left(
            s_\text{real}(x)
            -
            s_\text{fake}(x)
        \right)
        \frac{\mathrm{d}G_\theta}{\mathrm{d}\theta}
    \right].
\label{eq:dmd_base}
\end{equation}
For readability, we suppress the diffusion timestep and write $s(x)$ for the score evaluated on a noised version of $x$, \ie $x_\tau=(1-\tau) x+\tau\cdot\epsilon$ where $\epsilon\sim\mathcal{N}(0,I)$, and timestep is $\tau \in [0,1]$.

While DMD was introduced for distillation of few-step text-to-image models \cite{yin2024one,yin2024improved,bandyopadhyay2025sd3}, it has recently been applied to distilling text-to-video and image-to-video models, including causal and frame-wise generators~\citep{yin2025causvid,zhang2025fast,huang2025selfforcing,zhu2026causal,zhao2026causal++,zheng2026causal,feng2026one}. In these applications, the distillation strategy largely follows the T2I setting: the student generates a full sample, and the teacher provides a score over the corresponding full video clip.

\noindent \textbf{Autoregressive Video Generation.} We represent an image-to-video latent sequence as $x \equiv \mathbf{x}_{0:T} = (\mathbf{x}_0,\ldots,\mathbf{x}_T)$, where $\mathbf{x}_0=\mathcal{I}_0$ is the fixed input-image latent and $\mathbf{x}_{1:T}$ are the generated video latents. The sequence $c_{1:T}$ represents per-frame conditioning signals such as camera motion or user controls for the generated frames. Standard video diffusion models (Diffusion Transformers) generate a full video clip jointly, typically using bidirectional temporal attention over all frames. This allows each latent frame to depend on both past and future frames during denoising, but the cost of full attention grows quadratically with the number of frames. As a result, extending such models to long or interactive generation is expensive. Autoregressive video generation instead decomposes the video distribution over time \cite{gao2024ca2,yin2025causvid,huang2025selfforcing}. In a chunk-wise autoregressive model, we partition the generated latents and their controls into $N$ aligned blocks $\text{B}_{1:N}$ and $C_{1:N}$, each containing $K$ frames. The model factorizes as
\begin{equation}
    p_\theta(\text{B}_{1:N}\mid \mathcal{I}_0,C_{1:N})
    =
    \prod_{i=1}^{N}
    p_\theta(\text{B}_i\mid \mathcal{I}_0,\text{B}_{<i},C_{\leq i}).
\end{equation}

Causality is enforced across generated blocks, while frames within the current block may interact bidirectionally. We treat the input image $\mathcal{I}_0$ as part of the ordinary causal history rather than something that is permanently retained beyond local attention windows.
Setting $K=1$ recovers frame-wise autoregressive generation \cite{zhao2026causal++,feng2026one}, whose factorization is
\begin{equation}
    p_\theta(\mathbf{x}_{1:T}\mid \mathbf{x}_0=\mathcal{I}_0,c_{1:T})
    =
    \prod_{t=1}^{T}
    p_\theta(
        \mathbf{x}_t
        \mid
        \mathbf{x}_{<t},
        c_{\leq t}
    ).
\label{eq:frame_ar}
\end{equation}

\noindent \textbf{Diffusion Forcing.} Diffusion Forcing~\cite{chen2024diffusion} trains causal sequence models by independently corrupting frames and denoising each target under a noisy history. By varying the corruption applied to past frames, it exposes the model to imperfect context during training rather than conditioning only on clean data, improving robustness when generated frames are reused as context at inference time. We use this objective to pre-train the causal teacher used for Context-Matched Distillation.

\section{Methodology}

We introduce \textbf{Context-Matched Distillation (CMD)} to distill a multi-step causal teacher into a few-step causal student for autoregressive video generation. We begin by fine-tuning a pretrained bidirectional teacher into multi-step causal and block-causal variants. The student models are initialized from the teacher weights and trained by scoring their generated samples against the teacher distribution. Specifically, the student is rolled out on-policy to produce generated targets, while the causal teacher scores each target under the same temporal information boundary as the student, without access to future frames or controls. Building on base CMD, \textbf{Prefix Scoring} replaces the noised scoring history for the current frame with the cached student-generated prefix used to generate it.
A block-causal attention mask pairs every target with its corresponding prefix while allowing all targets to be scored in a single teacher pass. We further introduce \textbf{Prefix Corruption} to reduce sensitivity to artifacts in a weak student's generated prefixes. \Cref{sec:motivation} motivates CMD by identifying the information-set mismatch between bidirectional teacher scoring and causal student generation. \Cref{sec:cmd} presents the core distillation method, including causal-teacher pre-training, Prefix Scoring, efficient block-causal scoring, and Prefix Corruption. Finally, \cref{sec:long_video} and \cref{sec:relative_cameras} present natural extensions of CMD to long-video generation and camera control, respectively, with minimal changes to its formulation.

\subsection{Motivation}
\label{sec:motivation}

\noindent \textbf{Conditioning mismatch in causal training.} We discuss the formulation for frame-wise causal video generation pipelines in \cref{eq:frame_ar}. This factorization makes direct DMD supervision from a bidirectional teacher problematic \cite{zhu2026causal}, where $s_{\text{real}}(\mathbf{x}_{1:T}) = \nabla_{\mathbf{x}_{1:T}}\log p_\text{real}(\mathbf{x}_{1:T}\mid \mathbf{x}_0=\mathcal{I}_0,c_{1:T})$. The fake sample in \cref{eq:dmd_base} is $\mathbf{x}_{1:T}$, but each frame $\mathbf{x}_t$, $t=1,\ldots,T$, is generated using only the causal information set $(\mathbf{x}_{<t},c_{\leq t})$. In contrast, a bidirectional teacher provides $s_\text{real}(\mathbf{x}_{1:T})$ by scoring the full video in its native full-clip denoising setting. Its supervision for frame $t$ can therefore depend on information that is unavailable when the causal student generates that frame, including future latents $\mathbf{x}_{>t}$ and future conditioning $c_{>t}$. Thus, although the DMD update is applied to a student sample, the teacher score is computed under a different information set from the one used by the student at inference time.

\begin{figure}[t]
    \centering
    \includegraphics[width=\linewidth]{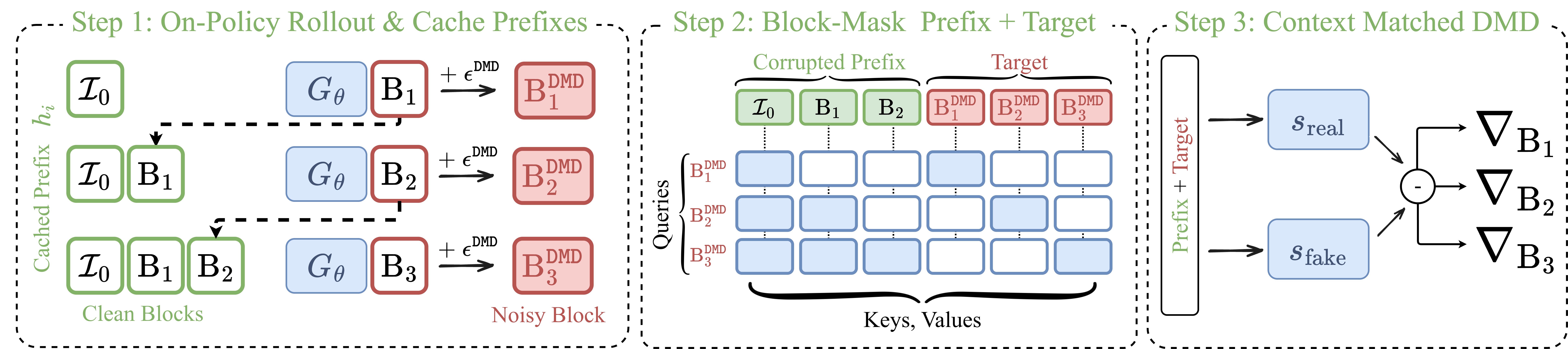}
    \caption{\textbf{Efficient Prefix Scoring.} The causal student first generates an on-policy rollout and caches the clean prefix used to produce each block. Corrupted prefixes and their corresponding noised DMD targets are then packed under a block-causal attention mask, allowing all targets to be scored in parallel. Finally, the real and fake scores are evaluated under the same prefix--target context to form the context-matched DMD update.}
    \label{fig:main_diag}
\end{figure}

Context-Matched Distillation (CMD) addresses this conditioning mismatch by replacing the bidirectional full-clip teacher score with a score from a causal teacher queried under the same information set available to the student. For $t=1,\ldots,T$, we use the causal information set $h_t = (\mathbf{x}_{<t}, c_{\leq t})$, where $\mathbf{x}_{<t}$ corresponds to past frames, including image conditioning $\mathbf{x}_0=\mathcal{I}_0$. The teacher score used in DMD for frame $t$ can be written as $s_{\mathrm{real},t}(\mathbf{x}_t,h_t) = \nabla_{\mathbf{x}_t} \log p_\text{real}(\mathbf{x}_t \mid h_t)$
rather than as the $t$-th component of a bidirectional full-clip score (see \cref{fig:concept}). The student score is evaluated under the same causal information boundary, yielding a frame-level score difference
$s_{\mathrm{real},t}(\mathbf{x}_t,h_t)
-
s_{\mathrm{fake},t}(\mathbf{x}_t,h_t)$.
This preserves the standard DMD score-difference update form while aligning the teacher supervision with the causal information available during autoregressive inference. The resulting information-set alignment and efficient joint scoring procedure are summarized in \cref{fig:main_diag}.

\subsection{Context-Matched Distillation}
\label{sec:cmd}

\noindent \textbf{Pre-training Causal Teacher.}
As discussed in \cref{sec:motivation}, CMD requires teacher scoring to follow the same causal information boundary as student generation. Towards this, we pre-train a causal multi-step teacher with a standard flow-matching objective to perform generation without bidirectional context.
Given a real video latent sequence $\mathbf{x}_{0:T}$ with $\mathbf{x}_0=\mathcal{I}_0$, we keep the input image clean and independently corrupt each input frame with noise at random timesteps. A block mask restricts target frame $t$ to $h_t=(\mathcal{I}_0,\tilde{\mathbf{x}}_{1:t-1},c_{1:t})$, where $\tilde{\mathbf{x}}_{1:t-1}$ denotes preceding frames corrupted independently. The teacher is trained with the standard flow-matching loss
\begin{equation}
 \mathcal{L}_{\mathrm{DF}} = \mathbb{E}\left\|v_t -\eta_\phi(\mathbf{x}_{t,\tau},\tau,h_t)\right\|_2^2,
\end{equation}
where $v_t = (\epsilon-\mathbf{x}_t)$. Following Diffusion Forcing \cite{chen2024diffusion}, this improves robustness to noisy context from the past. The learned flow field $\eta_\phi(\mathbf{x}_{t,\tau},\tau,h_t)$ can be used to score the causal student pipeline as $s_\text{real}(\mathbf{x}_{t,\tau},\tau,h_t)$, following \cite{yin2024one}.

\noindent \textbf{Causal Student Distillation.} We use the pretrained causal teacher $\eta_\phi$ for initializing the few-step student $G_\theta$ to carry forward the teacher's causal inductive bias. \textbf{Notably, our formulation does not require expensive student initialization through ODE \cite{zhuang2026self,yin2025causvid} or Consistency Distillation \cite{zhu2026causal,zhao2026causal++} based pre-training.} During distillation, the student is rolled out exactly as it would be during inference, following backward-simulation \cite{yin2024improved} in Self-Forcing \cite{huang2025selfforcing}. Specifically, we set $\hat{\mathbf{x}}_0=\mathcal{I}_0$ and use the student $G_\theta$ to generate $\hat{\mathbf{x}}_{1:T}$. For $t=1,\ldots,T$, latent frames are generated with velocities $\hat{v}_{t,\tau}=G_\theta(\hat{\mathbf{x}}_{t,\tau},\tau,h_t)$, where $h_t$ corresponds to past-frame latents detached from gradient computation.
For base CMD, each student-generated target $\hat{\mathbf{x}}_t$ is noised to obtain $\hat{\mathbf{x}}_{t,\tau}^{\texttt{DMD}}$. These noised targets are arranged temporally and scored with the causal teacher and fake-score network using a block-causal attention mask. The scoring context for target $t$ is
$
h_t^{\texttt{CMD}}
=(
    \mathcal{I}_0,
    \{\hat{\mathbf{x}}_{i,\tau}^{\texttt{DMD}}\}_{i=1}^{t-1},
    c_{1:t}
),
$
where $\hat{\mathbf{x}}_{<t,\tau}^{\texttt{DMD}}$ contains the preceding noised DMD targets. The causal attention mask prevents the score for $\hat{\mathbf{x}}_{t,\tau}^{\texttt{DMD}}$ from accessing future frames or controls. Base CMD provides causal teacher supervision matched to the temporal information boundary of student generation.

\noindent \textbf{Prefix Scoring.}
Although base CMD matches the temporal information boundary, its scoring history consists of preceding noised DMD targets rather than the realized history under which the student generated the current target. In practice, we instantiate CMD with Prefix Scoring. During the on-policy student rollout, we cache the generated history $\hat{\mathbf{x}}_{<t}$ and controls $c_{\leq t}$ that condition each target. For DMD supervision, frame $\hat{\mathbf{x}}_t$ generated by the student is noised at timestep $\tau$ with $\epsilon^{\texttt{DMD}}$ to obtain $\hat{\mathbf{x}}_{t,\tau}^{\texttt{DMD}}$. The frozen causal teacher $\eta_\phi$ and the fake-score network evaluate this noised target using the cached causal history $h_t=(\mathcal{I}_0, \hat{\mathbf{x}}_{1:t-1}, c_{1:t})$, yielding $s_\text{real}(\hat{\mathbf{x}}_{t,\tau}^{\texttt{DMD}},\tau,h_t)$ and $s_\text{fake}(\hat{\mathbf{x}}_{t, \tau}^{\texttt{DMD}},\tau,h_t)$, respectively, for the update in \cref{eq:dmd_base}. Rather than using previous noised DMD targets as causal context, Prefix Scoring therefore matches the scoring context to the context under which the target was generated.
Naively, Prefix Scoring would require a separate teacher pass for each frame. Instead, the teacher receives the input image and student-generated frames as conditioning memory, and uses a block-causal mask which lets each scoring query attend only to its causal prefix $\hat{\mathbf{x}}_{<t}$ and controls $c_{\leq t}$, allowing all generated frames to be scored jointly in one pass, akin to Teacher Forcing \cite{gao2024ca2}.

\noindent \textbf{Prefix Corruption.} Autoregressive video methods have addressed rollout error through self-generated histories \cite{huang2025selfforcing}, self-resampling that simulates imperfect histories \cite{guo2025end}, anti-drifting context construction \cite{zhang2026frame}, and gradients through historical context representations \cite{zhuang2026self}. Complementary to these, a naive teacher scoring on the exact generated prefix is not ideal, as the student can drift heavily right after initialization in few-step settings. This can dominate training when the causal teacher score is conditioned on heavily drifted, out-of-distribution artifacts. To prevent student drift from dominating score quality, we introduce prefix corruption, where the conditioning prefix is corrupted with Gaussian noise that reduces teacher focus on common drift artifacts such as color and texture. Specifically, we corrupt the prefix as:
\begin{equation}
\label{eq:dmd_prefix_corruption}
    \tilde{h}_t^{(\rho)}
    =
    \left(
        \mathcal{I}_0,
        \mathcal{C}_{\rho}(\hat{\mathbf{x}}_{1:t-1}),
        c_{1:t}
    \right),
    \qquad
    \mathcal{C}_{\rho}(\hat{\mathbf{x}}_{1:t-1})
    =
    (1-\rho)\hat{\mathbf{x}}_{1:t-1}
    +
    \rho \cdot \epsilon,
    \quad
    \epsilon \sim \mathcal{N}(0,I).
\end{equation}
Here $\rho$ is a corruption hyperparameter and is set to a small value by default. The Prefix Scoring DMD objective can now be written as:
\begin{equation}
    \nabla_\theta \mathcal{L}_{\mathrm{DMD}}
    =
    -\mathbb{E}_{\hat{\mathbf{x}}_{t,\tau}^{\texttt{DMD}},\,\tau}
    \left[
        \left(
            s_\text{real}(\hat{\mathbf{x}}_{t,\tau}^{\texttt{DMD}},\tau,\tilde{h}_t^{(\rho)})
            -
            s_\text{fake}(\hat{\mathbf{x}}_{t,\tau}^{\texttt{DMD}},\tau,\tilde{h}_t^{(\rho)})
        \right)
        \frac{\mathrm{d}G_\theta}{\mathrm{d}\theta}
    \right].
\label{eq:dmd_prefix_objective}
\end{equation}

Separate from noising in naive causal DMD, Prefix Corruption is a controlled relaxation of Prefix Scoring to stabilize distillation, and perturbs the on-policy prefix rather than the base CMD scoring sequence. The prefix is additionally corrupted using a dedicated noise level, independent of randomized DMD timesteps for scoring. This controls reliability of the generated context for the teacher without changing the target-side DMD.

\subsection{Long-Video Prefixes}
\label{sec:long_video}

\begin{figure}[t]
    \centering
    \includegraphics[width=\linewidth]{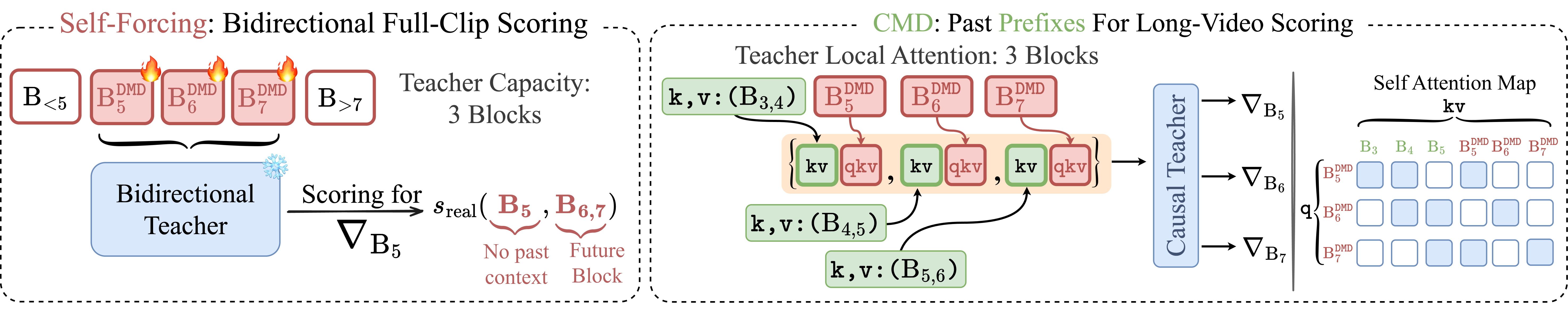}
    \caption{\textbf{Prefix Scoring for long-video distillation.} A fixed-capacity bidirectional teacher treats each local clip as self-contained, omitting earlier rollout context while allowing access to future blocks. Prefix Scoring instead evaluates all targets in one causal teacher pass under their corresponding bounded rollout prefixes.}
    \label{fig:long_video_rollout}
\end{figure}

Causal long-video models manage their history through rolling windows and sink frames \cite{yang2025longlive,liu2025rolling,yesiltepe2026infinity}, context packing or cache compression \cite{zhang2026frame,yi2025deep,mao2026packforcing}, or learned long-term memory \cite{yu2025videossm,xiao2026worldmem}. We use a local attention window for both the student and the causal teacher, as illustrated in \cref{fig:long_video_rollout}. Let $M$ denote the number of preceding frames retained by the student. For $t>M$, the student generates frame $t$ under the bounded causal context $h_t^M=(\hat{\mathbf{x}}_{t-M:t-1},c_{t-M:t})$ in $G_\theta(\hat{\mathbf{x}}_{t,\tau},\tau,h_t^M)$. Under the same memory budget, a bidirectional teacher instead scores a bounded clip such as $\hat{\mathbf{x}}_{t-M:t}$ as a self-contained sample. This omits the pre-clip context that influenced the clip's earliest frames during the student rollout and allows those frames to attend to later frames in the clip. The resulting score therefore does not match the student's causal factorization.

In contrast, CMD replaces the bidirectional local-clip score with a causal teacher that prevents each target from accessing later frames in the scoring window. With Prefix Scoring, the causal teacher additionally evaluates each target under exactly the same bounded context as the student, \ie $h_t^M=(\hat{\mathbf{x}}_{t-M:t-1},c_{t-M:t})$. In practice, we train $(RT+1)$-frame videos using $R$ contiguous rollouts of $T$ generated frames, with $\mathcal{I}_0$ included only in the first rollout. After each rollout, its generated frames are cached as context for the next. Consequently, the context of a frame near a rollout boundary may include frames from the preceding rollout. These cached frames provide the local attention context and, when applicable, the sink tokens for both the student and the causal teacher.

For long-video training, we replace the fixed prefix-corruption level $\rho$ in \cref{eq:dmd_prefix_corruption} with a frame-dependent schedule $\rho_t$ over the full rollout. Early frames, before substantial drift has accumulated, receive weak prefix corruption, whereas later frames receive stronger corruption as rollout error grows. Applying the score-distillation loss to every generated frame across the $R$ rollouts therefore supervises the full causal trajectory while requiring attention only over a bounded prefix for each target. Longer target videos add rollout chunks to be scored without expanding the teacher's attention window, allowing the same local causal formulation to scale to arbitrarily long videos.

\subsection{Camera-Conditioned Causal Distillation}
\label{sec:relative_cameras}

Existing video models encode pose sequences through ray maps, camera adapters, or pose-aware attention \cite{he2024cameractrl,wang2024motionctrl,li2026prope,he2025cameractrl}, and recent autoregressive systems support interactive camera updates during streaming generation \cite{xu2026realcam,team2026advancing}. Prior work has used cameras relative to the previous frame \cite{team2026advancing}, while relative camera representations have shown improvements in novel-view synthesis quality over absolute alternatives \cite{li2026prope}. We adopt this parameterization of frame-relative cameras for training our teacher and student pipelines, embedding the camera as the motion of the current view relative to a generated view. Let $E_t \in SE(3)$ denote the camera-to-world pose for frame $t=0,\ldots,T$ and $K_t$ the corresponding intrinsics.  We convert the trajectory into frame-relative increments
\begin{equation}
    \Delta E_t =
    \begin{cases}
        I, & t=0,\\
        E_{t-1}^{-1} E_t, & t>0,
    \end{cases}
\label{eq:relative_camera}
\end{equation}
and use these increments, rather than the absolute pose sequence, as the camera condition.  For each latent-frame location $(u,v)$, we unproject a ray using $K_t$ and transform it by $\Delta E_t$ to obtain a spatial ray embedding
\begin{equation}
    c_t = \Phi_t(u,v) = \phi(\Delta E_t, K_t, u, v),
\end{equation}
where $\phi$ is implemented as a ray map \cite{team2026advancing}.  This representation is spatially aligned with the video tokens and preserves the local geometry of the commanded camera motion, while removing dependence on an arbitrary global coordinate frame.   The key distinction is how the camera representation is used by the teacher score.  In full-clip distillation, even a relative camera trajectory can be provided to a bidirectional teacher as a complete sequence.  The score for frame $t$ may then depend on future camera updates $\Phi_{>t}$, allowing the
teacher to explain the current frame using motion that is unavailable to the autoregressive student at inference time.  This creates the same information-set mismatch as future-frame lookahead: the student must choose $\mathbf{x}_t$ from $(\mathbf{x}_{<t},c_{\leq t})$, while the teacher score is computed with access to the whole trajectory. 
CMD prevents this mismatch by restricting the teacher score for frame $t$ to visual context and camera embeddings available up to that frame. With Prefix Scoring, this visual context is the cached generated prefix under which the target was produced.

 \section{Experiments}

\begin{table}[t]
    \centering
    \small
    \caption{Short-video comparison of CMD and autoregressive baselines on VBench-I2V. Q: Quality Score; IS/IB: I2V Subject/Background Consistency; CM: Camera Motion; SC/BC: Subject/Background Consistency; TF: Temporal Flickering; MS: Motion Smoothness; AQ/IQ: Aesthetic/Imaging Quality; DD: Dynamic Degree.}
    \label{tab:vbench}
    \setlength{\tabcolsep}{2pt}
    \begin{tabularx}{\linewidth}{@{}lY*{3}{Y}|*{10}{Y}@{}}
        \toprule
        \multirow{2}{*}{Method} & \multirow{2}{*}{\shortstack{Chunk\\Size}} & \multicolumn{3}{c|}{Aggregate} & \multicolumn{3}{c}{I2V}
        & \multicolumn{4}{c}{Consistency} & \multicolumn{3}{c}{Quality} \\
        \cmidrule(lr){3-5} \cmidrule(lr){6-8} \cmidrule(lr){9-12} \cmidrule(lr){13-15}
        & & Total & I2V & Q & IS & IB & CM & SC & BC & TF & MS & AQ & IQ & DD \\
        \midrule
        CausVid~\cite{yin2025causvid}
        & 3 & 80.77 & 84.06 & 77.48 & 89.05 & 90.99 & 15.57
        & 93.45 & 95.35 & 98.14 & 98.67 & 63.13 & 70.43 & 20.57 \\
        Self-Forcing~\cite{huang2025selfforcing}
        & 3 & 85.69 & 92.02 & 79.35 & 95.74 & 96.76 & 26.13
        & 96.82 & 96.80 & 98.04 & 99.11 & 62.29 & 72.31 & 24.15 \\
        LongLive~\cite{yang2025longlive}
        & 3 & 84.93 & 91.22 & 78.64 & 95.03 & 96.22 & 24.90
        & 96.72 & 96.97 & 98.43 & 99.24 & 62.76 & 72.04 & 14.80 \\
        Rolling Forcing~\cite{liu2025rolling}
        & 3 & 85.54 & 92.61 & 78.46 & 96.24 & 97.28 & 25.69
        & 96.37 & 96.89 & 98.04 & 98.90 & 62.36 & 72.01 & 17.07 \\
        Context Forcing~\cite{chen2026contextforcing}
        & 3 & 85.41 & 90.66 & 80.16 & 94.47 & 95.63 & 27.76
        & 94.75 & 95.90 & 97.80 & 98.74 & 62.79 & 72.11 & 42.28 \\
        LingBot-World~\cite{team2026advancing}
        & 3 & 86.86 & 93.76 & 79.96 & 96.00 & 98.04 & 42.54
        & 91.96 & 95.16 & 96.06 & 97.37 & 62.77 & 68.96 & 64.39 \\
        Causal Forcing~\cite{zhu2026causal}
        & 1 & 87.63 & 93.70 & \bestscore{81.56} & 96.57 & 98.04 & 34.47
        & 91.22 & 93.50 & 94.74 & 98.15 & 59.42 & 70.13 & 87.15 \\
        Causal Forcing++~\cite{zhao2026causal++}
        & 1 & 87.35 & 95.36 & 79.35 & 98.63 & 99.35 & 27.34
        & 96.35 & 97.41 & 98.70 & 99.19 & 62.82 & 71.48 & 23.66 \\
        \midrule
        \textbf{Ours}
        & 1 & \secondscore{88.46} & \secondscore{96.44} & \secondscore{80.47}
        & 98.12 & 98.79 & 63.64 & 95.77 & 97.36 & 97.13 & 98.14
        & 60.62 & 71.91 & 48.21 \\
        \textbf{Ours}
        & 4 & \bestscore{88.47} & \bestscore{96.54} & 80.40
        & 97.64 & 98.44 & 76.12 & 94.77 & 96.72 & 96.71 & 97.99
        & 60.82 & 71.51 & 52.93 \\
        \bottomrule
    \end{tabularx}
\end{table}

\subsection{Training}
We fine-tune Cosmos-Predict2.5-2B \cite{ali2025world} for autoregressive I2V generation with generated and curated videos \cite{wan2025wan} for non-camera controlled video generation and with DL3DV \cite{ling2024dl3dv} for camera-controlled video generation. We generate video captions for all generated videos with Qwen3-VL-8B-Instruct \cite{qwen3technicalreport}. This data combination contains diverse camera motion coupled with dynamic environments involving people and commonplace objects. We first train the teacher model on generated data as a causal multi-step video generator and then distill it into a few-step generator. For non-camera controlled video generation, the chunk-1/chunk-4 teachers are both trained for $\sim$8K iterations. Short-video chunk-1/chunk-4 students are trained for $\sim$3.2K/5.1K iterations while long-video students are fine-tuned from short-video checkpoints for an additional $\sim$0.8K/0.9K iterations respectively. For camera-controlled I2V, we train the chunk-1/chunk-4 teachers and students for $\sim$11K/8K and $\sim$3K/0.9K iterations, respectively.

\noindent \textbf{Baselines: Autoregressive I2V.}
Among native few-step block-causal I2V systems, we compare with LingBot-World~\cite{team2026advancing}, which trains a bidirectional teacher model as a Mixture-of-Experts-style DiT and distills it into a causal student with DMD for low-latency interactive generation. In addition to native I2V, we compare with frame-causal and block-causal T2V pipelines adapted to I2V by replacing the first generated frame with the input image:  (i) CausVid~\cite{yin2025causvid} trains a causal student model from a bidirectional teacher with DMD, where the student is initialized by ODE matching against the teacher's trajectory.  (ii) Self-Forcing~\cite{huang2025selfforcing} builds upon CausVid's~\cite{yin2025causvid} distillation recipe, but trains on samples rolled out by the student itself rather than only clean video data, reducing exposure bias from generated-history conditioning at inference time.  (iii) LongLive~\cite{yang2025longlive} targets real-time long-video generation using a bounded rolling context.  (iv) Rolling Forcing~\cite{liu2025rolling} performs autoregressive generation with a rolling denoising window, where multiple future frames are kept at different noise levels and shifted forward as frames are emitted; it also uses attention sinks to preserve early-frame context during long rollouts.  (v) Context Forcing~\cite{chen2026contextforcing} uses long-context teacher supervision to improve consistency across autoregressive rollouts.  (vi) Causal Forcing~\cite{zhu2026causal} modifies the ODE-matching initialization in CausVid~\cite{yin2025causvid} by matching the student against a causal teacher instead of the original bidirectional Wan~\cite{wan2025wan} teacher.  (vii) Causal Forcing++~\cite{zhao2026causal++} keeps the causal teacher from Causal Forcing~\cite{zhu2026causal} but replaces ODE-matching initialization with causal consistency distillation, after which the student is further optimized with DMD.

\begin{table}[t]
    \centering
    \small
    \caption{Long-video comparison of CMD and autoregressive baselines on SANA-WM~\cite{zhu2026sanawm}. Q/S: Quality/Semantic Scores; SC/BC: Subject/Background Consistency; TF: Temporal Flickering; MS: Motion Smoothness; DD: Dynamic Degree; AQ/IQ: Aesthetic/Imaging Quality.}
    \label{tab:long_video_vbench}
    \setlength{\tabcolsep}{2pt}
    \begin{tabularx}{\linewidth}{@{}lY*{3}{Y}|*{7}{Y}@{}}
        \toprule
        \multirow{2}{*}{Method} & \multirow{2}{*}{\shortstack{Chunk\\Size}} & \multicolumn{3}{c|}{Aggregate} & \multicolumn{4}{c}{Consistency}
        & \multicolumn{3}{c}{Quality} \\
        \cmidrule(lr){3-5} \cmidrule(lr){6-9} \cmidrule(lr){10-12}
        & & Q & S & Total & SC & BC & TF & MS & DD & AQ & IQ \\
        \midrule
        LingBot-World~\cite{team2026advancing}
        & 3 & 80.23 & 24.20 & 69.03
        & 91.00 & 93.82 & 97.35 & 98.11 & 51.25 & 60.20 & 67.37 \\
        LongLive~\cite{yang2025longlive}
        & 3 & 80.04 & 23.86 & 68.80
        & 95.43 & 95.08 & 98.38 & 99.16 & 8.75 & 61.20 & 73.07 \\
        Rolling Forcing~\cite{liu2025rolling}
        & 3 & 80.04 & 23.66 & 68.77
        & 96.09 & 95.41 & 98.60 & 99.13 & 3.75 & 61.63 & 73.48 \\
        Context Forcing~\cite{chen2026contextforcing}
        & 3 & 80.72 & 23.38 & 69.25
        & 91.72 & 93.45 & 98.06 & 98.95 & 32.50 & 62.03 & 72.92 \\
        \midrule
        \textbf{Ours}
        & 1 & \bestscore{81.39} & \bestscore{24.51} & \bestscore{70.02}
        & 88.59 & 91.93 & 96.24 & 98.13 & 67.50 & 60.70 & 74.58 \\
        \textbf{Ours}
        & 4 & \secondscore{80.84} & \secondscore{24.50} & \secondscore{69.57}
        & 86.97 & 91.34 & 95.83 & 97.84 & 77.50 & 60.50 & 71.02 \\
        \bottomrule
    \end{tabularx}
\end{table}

\begin{table}[H]
    \centering
    \small
    \caption{Camera-controlled video generation on the Simple and Hard SANA-WM~\cite{zhu2026sanawm} benchmark splits.}
    \label{tab:camera_accuracy}
    \setlength{\tabcolsep}{2pt}
    \begin{tabularx}{\linewidth}{@{}lYl*{3}{Y}|*{3}{Y}@{}}
        \toprule
        \multirow{2}{*}{Method} & \multirow{2}{*}{Chunk Size} & \multirow{2}{*}{Split}
        & \multicolumn{3}{c}{SANA-WM~\cite{zhu2026sanawm} VBench Scores} & \multicolumn{3}{c}{Camera Error} \\
        \cmidrule(lr){4-6} \cmidrule(lr){7-9}
        & & & Quality $\uparrow$ & Semantic $\uparrow$ & Total $\uparrow$
               & Rot. $\downarrow$ & Trans. $\downarrow$ & CamMC $\downarrow$ \\
        \midrule
        \multirow{2}{*}{HY-WorldPlay~\cite{sun2025worldplay}}
               & \multirow{2}{*}{4} & Simple & 0.8178 & 0.1786 & 0.6900
               & 2.5840 & 0.1639 & 0.1866 \\
               & & Hard & \secondscore{0.8215} & 0.1740 & \secondscore{0.6920}
               & 6.5991 & 0.2483 & 0.3225 \\
        \midrule
        \multirow{2}{*}{LingBot-World~\cite{team2026advancing}}
               & \multirow{2}{*}{3} & Simple & 0.8173 & \secondscore{0.1841} & 0.6906
               & 4.7122 & 0.1562 & 0.2194 \\
               & & Hard & 0.8144 & \secondscore{0.1792} & 0.6873
               & 6.2291 & 0.1774 & 0.2618 \\
        \midrule
        \multirow{2}{*}{SANA-WM~\cite{zhu2026sanawm}}
               & \multirow{2}{*}{3} & Simple & \bestscore{0.8355} & 0.1061 & 0.6896
               & \secondscore{1.2104} & 0.1316 & 0.1416 \\
               & & Hard & \bestscore{0.8263} & 0.1073 & 0.6825
               & \secondscore{2.0453} & 0.1822 & 0.1982 \\
        \midrule
        \multirow{2}{*}{minWM~\cite{zhao2026minwm}}
               & \multirow{2}{*}{4} & Simple & 0.8102 & 0.1826 & 0.6847
               & 2.8637 & 0.2174 & 0.2388 \\
               & & Hard & 0.8167 & 0.1775 & 0.6889
               & 6.9128 & 0.2811 & 0.3538 \\
        \midrule
        \multirow{2}{*}{\textbf{Ours}}
               & \multirow{2}{*}{1} & Simple & \secondscore{0.8260} & \bestscore{0.2413} & \bestscore{0.7091}
               & 1.3675 & \secondscore{0.1195} & \secondscore{0.1312} \\
               & & Hard & 0.8184 & \bestscore{0.2423} & \bestscore{0.7032}
               & 2.5607 & \secondscore{0.1447} & \secondscore{0.1698} \\
        \midrule
        \multirow{2}{*}{\textbf{Ours}}
               & \multirow{2}{*}{4} & Simple & 0.8195 & 0.1810 & \secondscore{0.6918}
               & \bestscore{0.8601} & \bestscore{0.0981} & \bestscore{0.1027} \\
               & & Hard & 0.8139 & 0.1751 & 0.6862
               & \bestscore{1.2718} & \bestscore{0.1109} & \bestscore{0.1196} \\
        \bottomrule
    \end{tabularx}
\end{table}

\noindent \textbf{Baselines: Action Control I2V.}
Relevant action-controllable I2V world models include the following. (i) HY-WorldPlay~\cite{sun2025worldplay} uses reconstituted context memory for long-horizon streaming generation together with PRoPE camera conditioning and discrete action embeddings. (ii) LingBot-World~\cite{team2026advancing} injects camera ray maps and discrete actions through adaptive normalization, then distills the causal model for real-time rollout. (iii) SANA-WM~\cite{zhu2026sanawm} combines latent-frame UCPE with raw-frame camera ray maps for 6-DoF camera control. (iv) minWM~\cite{zhao2026minwm} provides a full-stack recipe for adding PRoPE camera conditioning to video backbones and converting them into few-step autoregressive world models.

\subsection{Quantitative Evaluation}

\noindent \textbf{Video Quality.}
For short-horizon I2V generation, we evaluate CMD with chunk sizes one and four on the VBench-I2V benchmark~\cite{huang2025vbench++}, including five fixed seeds per image--prompt pair for every method. We report the official normalized Total, I2V, and Quality scores together with the raw component scores in \cref{tab:vbench}. Our chunk-4 model achieves the best Total (88.47), I2V (96.54), and Camera Motion (76.12) scores, improving over the strongest baseline by 0.84, 1.18, and 33.58 percentage points, respectively. Our chunk-1 model ranks second on all three metrics, with scores of 88.46, 96.44, and 63.64.

\noindent \textbf{Long-Video Quality.}
For long-horizon evaluation, we use the 80 images and original prompts from the Simple split of the SANA-WM benchmark~\cite{zhu2026sanawm}. Each method is evaluated over 501 frames at 16 fps ($\sim$30 seconds) and $480{\times}832$ resolution without camera input. We report VBench scores~\cite{huang2025vbench++}, as in SANA-WM~\cite{zhu2026sanawm}, in \cref{tab:long_video_vbench}. Our chunk-1 model achieves the best score on every aggregate metric in the main comparison. Relative to the strongest external baseline in each column, it improves Q by 0.67 percentage points over Context Forcing~\cite{chen2026contextforcing}, S by 0.31 points over LingBot-World~\cite{team2026advancing}, and Total by 0.77 points over Context Forcing~\cite{chen2026contextforcing}. Our chunk-4 model ranks second on every aggregate metric, with Q, S, and Total scores of 80.84, 24.50, and 69.57, respectively, and has the highest raw Dynamic Degree (0.7750). LongLive \cite{yang2025longlive} and Rolling Forcing \cite{liu2025rolling} attain higher temporal-consistency scores but substantially lower Dynamic Degree (0.0875 and 0.0375, respectively).

\noindent \textbf{Camera-Controlled Evaluation.}
For camera-controlled evaluation, we use the Simple and Hard splits of the SANA-WM benchmark~\cite{zhu2026sanawm}. Each method generates eight 125-frame windows per scene, yielding 640 videos per split. Following SANA-WM~\cite{zhu2026sanawm}, we report Quality, Semantic, and Total scores together with rotation, translation, and Camera Matrix Consistency (CamMC) errors in \cref{tab:camera_accuracy}. To compute the camera errors, we stitch the windows, recover camera poses with $\pi^3$~\cite{wang2025pi}, and align the recovered trajectory to the target using Umeyama $\mathrm{Sim}(3)$~\cite{umeyama1991least}. Rotation error is the mean angular difference in degrees, translation error is the mean Euclidean distance, and CamMC is the mean Frobenius distance between the aligned $3{\times}4$ camera matrices; all three are lower-is-better. Our chunk-1 model achieves the best Semantic and Total scores on both splits, while our chunk-4 model achieves the lowest rotation, translation, and CamMC errors on both splits.

\begin{table}[H]
    \centering
    \small
    \caption{Ablations over short-video generation with CMD variants on VBench-I2V. PS denotes Prefix Scoring and PC$_k$ Prefix Corruption with $t_{\mathrm{prefix}}=k$; Full CMD uses PS with PC$_{256}$. Q: Quality Score; IS/IB: I2V Subject/Background Consistency; CM: Camera Motion; SC/BC: Subject/Background Consistency; TF: Temporal Flickering; MS: Motion Smoothness; AQ/IQ: Aesthetic/Imaging Quality; DD: Dynamic Degree.}
    \label{tab:short_video_ablation}
    \setlength{\tabcolsep}{2pt}
    \begin{tabularx}{\linewidth}{@{}l*{3}{Y}|*{10}{Y}@{}}
        \toprule
        \multirow{2}{*}{Method} & \multicolumn{3}{c|}{Aggregate} & \multicolumn{3}{c}{I2V}
        & \multicolumn{4}{c}{Consistency} & \multicolumn{3}{c}{Quality} \\
        \cmidrule(lr){2-4} \cmidrule(lr){5-7} \cmidrule(lr){8-11} \cmidrule(lr){12-14}
        & Total & I2V & Q & IS & IB & CM & SC & BC & TF & MS & AQ & IQ & DD \\
        \midrule
        Bidir. teacher
        & 82.65 & 90.81 & 74.48 & 93.90 & 95.74 & 36.23
        & 85.10 & 90.28 & 94.86 & 96.16 & 53.80 & 57.37 & 82.85 \\
        Base CMD
        & \secondscore{88.33} & \secondscore{96.40} & \secondscore{80.25}
        & 98.25 & 98.84 & 60.66 & 96.08 & 97.55 & 97.58 & 98.36
        & 61.02 & 71.89 & 42.36 \\
        + PS
        & 87.25 & 94.83 & 79.67 & 97.20 & 97.69 & 55.49
        & 92.47 & 94.33 & 94.79 & 97.07 & 58.53 & 70.83 & 69.02 \\
        + PS + PC$_{128}$
        & 88.04 & 96.18 & 79.90 & 98.20 & 98.86 & 56.15
        & 95.53 & 97.38 & 97.24 & 98.18 & 60.49 & 71.91 & 42.44 \\
        + PS + PC$_{512}$
        & 88.06 & 96.23 & 79.89 & 98.31 & 98.88 & 55.75
        & 96.28 & 97.67 & 97.62 & 98.37 & 61.06 & 71.98 & 37.15 \\
        \midrule
        \textbf{Full CMD} (PC$_{256}$)
        & \bestscore{88.46} & \bestscore{96.44} & \bestscore{80.47}
        & 98.12 & 98.79 & 63.64 & 95.77 & 97.36 & 97.13 & 98.14
        & 60.62 & 71.91 & 48.21 \\
        \bottomrule
    \end{tabularx}
\end{table}

\subsection{Ablative experiments}
\noindent \textbf{Short video generation.} We report ablations over short-video generation in \cref{tab:short_video_ablation}. We compare (i) \emph{Bidir. teacher}: the bidirectional teacher jointly scores the complete generated clip and can attend to both past and future frames; and (ii) \emph{Base CMD}: the teacher is causal, but previous noisy DMD frames provide its context instead of a separately cached prefix. We then ablate prefix corruption using the scheduler timestep $t_{\mathrm{prefix}}$, which determines $\rho$: (iii) \emph{Base CMD + Prefix Scoring ($t_{\mathrm{prefix}}=0$)}: the generated prefixes remain clean; (iv) \emph{Full CMD w. $t_{\mathrm{prefix}}=128$}: weaker noise is added to the prefixes; (v) \emph{Full CMD w. $t_{\mathrm{prefix}}=256$}: our default prefix-noise level is used; and (vi) \emph{Full CMD w. $t_{\mathrm{prefix}}=512$}: the prefixes receive stronger corruption. These four configurations otherwise use the same Prefix Scoring setup. Moving from bidirectional scoring to base CMD raises Total by 5.68 points and places base CMD second on all aggregate metrics. Compared with base CMD, our default prefix-conditioned model further improves Total by 0.13 points, Camera Motion by 2.98 points, and Dynamic Degree by 5.85 points. Clean prefixes do not automatically improve aggregate performance over Base CMD; prefix corruption is important, with $t_{\mathrm{prefix}}=256$ giving the best aggregate result and Camera Motion score, while weaker or stronger corruption gives lower Camera Motion scores.

\begin{table}[H]
    \centering
    \small
    \caption{Ablations over long-video generation with CMD variants on SANA-WM~\cite{zhu2026sanawm} benchmark. Full CMD includes Prefix Scoring and Prefix Corruption. Q/S: Quality/Semantic Scores; SC/BC: Subject/Background Consistency; TF: Temporal Flickering; MS: Motion Smoothness; DD: Dynamic Degree; AQ/IQ: Aesthetic/Imaging Quality.}
    \label{tab:long_video_ablation}
    \setlength{\tabcolsep}{2pt}
    \begin{tabularx}{\linewidth}{@{}l*{3}{Y}|*{7}{Y}@{}}
        \toprule
        \multirow{2}{*}{Method} & \multicolumn{3}{c|}{Aggregate} & \multicolumn{4}{c}{Consistency}
        & \multicolumn{3}{c}{Quality} \\
        \cmidrule(lr){2-4} \cmidrule(lr){5-8} \cmidrule(lr){9-11}
        & Q & S & Total & SC & BC & TF & MS & DD & AQ & IQ \\
        \midrule
        Bidir. teacher
        & 76.49 & \secondscore{23.93} & 65.98
        & 83.66 & 89.77 & 95.32 & 96.59
        & 67.50 & 54.45 & 65.44 \\
        Base CMD
        & \secondscore{80.75} & 23.86 & \secondscore{69.37}
        & 87.44 & 91.44 & 96.16 & 98.06
        & 70.00 & 59.12 & 73.20 \\
        \midrule
        \textbf{Full CMD}
        & \bestscore{81.39} & \bestscore{24.51} & \bestscore{70.02}
        & 88.59 & 91.93 & 96.24 & 98.13
        & 67.50 & 60.70 & 74.58 \\
        \bottomrule
    \end{tabularx}
\end{table}

\begin{table}[H]
    \centering
    \small
    \caption{Camera-control ablations on the SANA-WM benchmark~\cite{zhu2026sanawm} over camera representation and CMD variants. All variants use chunk size 1. Full CMD includes Prefix Scoring and Prefix Corruption.}
    \label{tab:camera_control_ablation}
    \setlength{\tabcolsep}{9pt}
    \begin{tabularx}{\linewidth}{@{}>{\raggedright\arraybackslash}Xcl*{3}{c}|*{3}{c}@{}}
        \toprule
        \multirow{2}{*}{Method}
        & \multirow{2}{*}{Camera}
        & \multirow{2}{*}{Split}
        & \multicolumn{3}{c|}{SANA-WM~\cite{zhu2026sanawm} VBench Scores $\uparrow$}
        & \multicolumn{3}{c}{Camera error $\downarrow$} \\
        \cmidrule(lr){4-6} \cmidrule(lr){7-9}
        & & & Quality & Semantic & Total & Rot. & Trans. & CamMC \\
        \midrule

        \multirow{2}{*}{Bidir. teacher}
        & \multirow{4}{*}{PRoPE}
        & Simple & 0.6794 & 0.1133 & 0.5662
        & 11.2383 & 0.5413 & 0.6726 \\
        & & Hard & 0.6779 & 0.1142 & 0.5651
        & 12.4233 & 0.5481 & 0.6885 \\
        \cmidrule(lr){1-1} \cmidrule(lr){3-9}
        \multirow{2}{*}{Full CMD}
        &
        & Simple & 0.7775 & 0.1736 & 0.6567
        & \bestscore{1.2546} & 0.4043 & 0.4083 \\
        & & Hard & 0.7750 & 0.1691 & 0.6538
        & 3.2393 & 0.4311 & 0.4543 \\

        \midrule

        \multirow{2}{*}{Bidir. teacher}
        & \multirow{6}{*}{Ray map}
        & Simple & 0.7998 & \secondscore{0.1878} & 0.6774
        & 5.7306 & 0.2858 & 0.3329 \\
        & & Hard & 0.7936 & \secondscore{0.1818} & 0.6712
        & 7.1274 & 0.3027 & 0.3733 \\
        \cmidrule(lr){1-1} \cmidrule(lr){3-9}
        \multirow{2}{*}{Base CMD}
        &
        & Simple & \secondscore{0.8214} & 0.1036 & \secondscore{0.6778}
        & 1.7265 & \secondscore{0.1264} & \secondscore{0.1416} \\
        & & Hard & \secondscore{0.8128} & 0.1054 & \secondscore{0.6713}
        & \bestscore{2.1606} & \bestscore{0.1257} & \bestscore{0.1468} \\
        \cmidrule(lr){1-1} \cmidrule(lr){3-9}
        \multirow{2}{*}{\textbf{Full CMD}}
        &
        & Simple & \bestscore{0.8260} & \bestscore{0.2413} & \bestscore{0.7091}
        & \secondscore{1.3675} & \bestscore{0.1195} & \bestscore{0.1312} \\
        & & Hard & \bestscore{0.8184} & \bestscore{0.2423} & \bestscore{0.7032}
        & \secondscore{2.5607} & \secondscore{0.1447} & \secondscore{0.1698} \\
        \bottomrule
    \end{tabularx}
\end{table}

\noindent \textbf{Long video generation.} We report ablations over long video generation in \cref{tab:long_video_ablation}. We compare (i) \emph{Bidir. teacher}: the score teacher processes each clip with bidirectional attention and no preceding rollout context; (ii) \emph{Base CMD}: the teacher is causal within each clip, but receives no cross-clip prefix; and (iii) \emph{Full CMD}: the causal teacher receives the same bounded cross-clip prefix as the student. All three causal students are evaluated using the same seed and 80 SANA-WM~\cite{zhu2026sanawm} prompts. Here, the prefix carries the student's bounded history across clip boundaries. Base CMD improves Total by 3.39 points over bidirectional scoring. Compared with Bidir. teacher, Full CMD improves Total by 4.04 points and Semantic by 0.58 points.
\begin{figure}[!t]
    \centering
    \includegraphics[width=\linewidth]{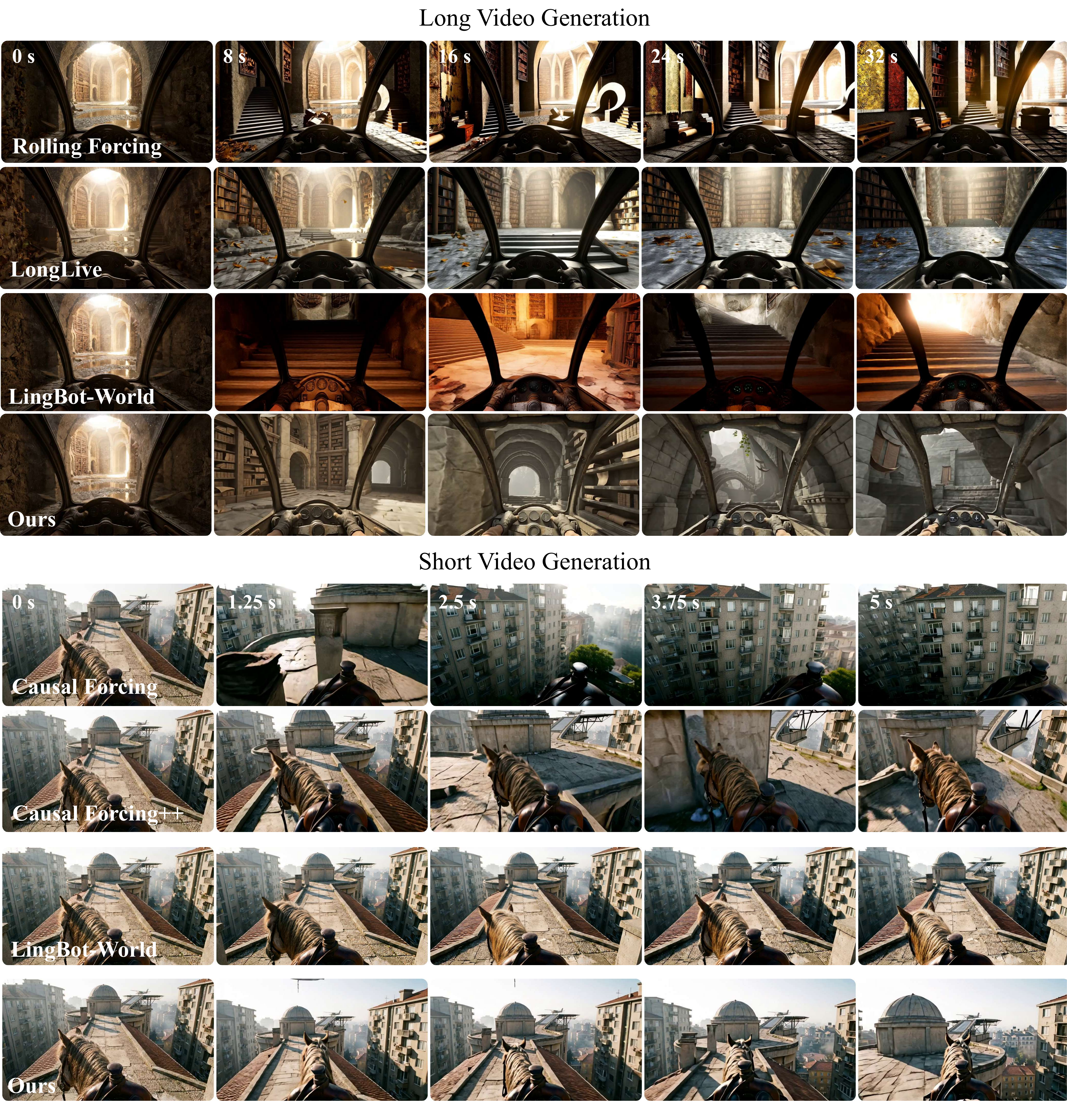}
    \vspace{-0.8cm}
    \caption{Qualitative comparisons for long-video (top) and short-video (bottom) generation.}
    \label{fig:qualitative_comparisons}
\end{figure}
\begin{figure}[!t]
    \centering
    \includegraphics[width=\linewidth]{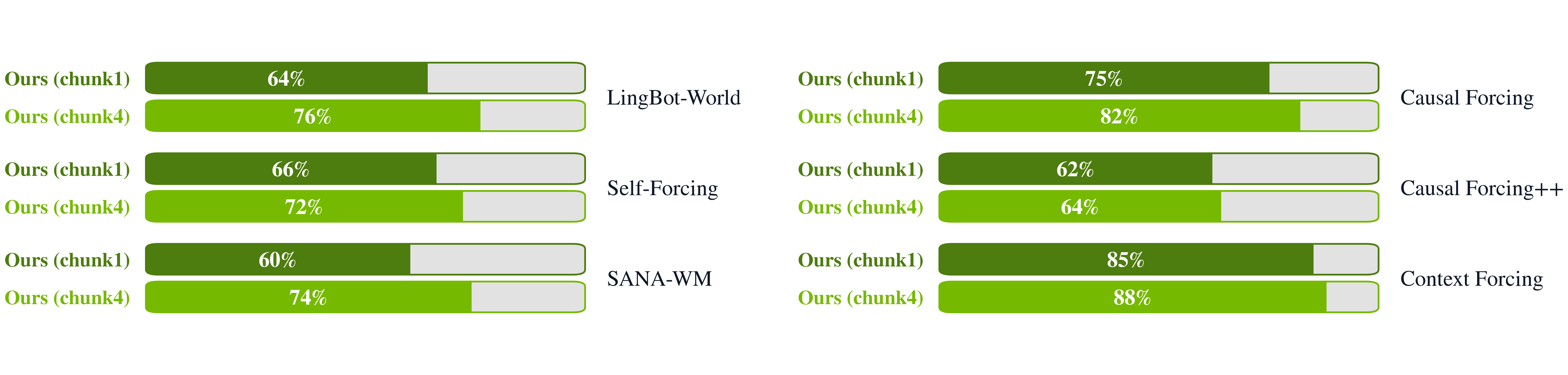}
    \vspace{-1cm}
    \caption{LLM preferences for our chunk-1 and chunk-4 models over 100 comparisons per model-baseline pair.}
    \label{fig:automated_preference}
\end{figure}

\noindent \textbf{Camera control.} We report chunk-1 camera-control ablations in \cref{tab:camera_control_ablation}. We compare (i) \emph{Bidir. teacher + PRoPE}: the teacher jointly scores the generated frames with bidirectional attention and PRoPE conditioning \cite{li2026prope}; (ii) \emph{Full CMD + PRoPE}: block-causal scoring is used with the same PRoPE conditioning; (iii) \emph{Bidir. teacher + Ray map}: the bidirectional teacher instead uses ray map conditioning; (iv) \emph{Base CMD + Ray map}: the teacher is causal but receives previous noisy DMD frames rather than a separately cached prefix; and (v) \emph{Full CMD + Ray map}: the causal teacher uses ray map conditioning and the cached student prefix. All variants use the same prompts and evaluation protocol. With PRoPE, our causal formulation improves every metric over bidirectional scoring. With ray map, Base CMD improves Quality by 0.0216/0.0192 over bidirectional scoring on Simple/Hard and reduces all camera errors. Adding the cached prefix then improves Total by a further 0.0313/0.0319, driven by Semantic gains of 0.1377/0.1369. It also further reduces all camera errors on the Simple split for the benchmark, while Base CMD retains lower camera errors on Hard. Compared with PRoPE, the full ray map model improves all SANA-WM VBench~\cite{zhu2026sanawm} metrics, translation, and CamMC on both splits.

\subsection{Qualitative Analysis}

\noindent \textbf{Visual comparisons.}
\Cref{fig:qualitative_comparisons} compares representative rollouts across both evaluation settings. In the long-video example, our model preserves the first-person vehicle structure while progressing through distinct indoor spaces over 32 seconds. The baselines show either less scene progression or larger discontinuities in the vehicle and surrounding structure. In the short-video example, our model keeps the horse coherent while moving it forward along the rooftop towards the dome. In contrast, the baselines either lose the horse, change its placement abruptly, or remain close to the initial view. These examples complement the aggregate evaluations by showing that our model maintains visual continuity while producing clear scene progression over both short and extended horizons.

\noindent \textbf{Automated pairwise preference.}
We use a blind, forced-choice LLM judge as our primary perceptual evaluation as it explicitly evaluates whether the requested action and camera motion are fulfilled. We use Gemini 3.1 Pro Preview with temperature $1$ and high reasoning effort. We sample 100 VBench-I2V examples with a fixed seed and compare our chunk-1 and chunk-4 models with every baseline using the same prompt and reference image. Within each pair, videos are randomly assigned to A/B without revealing method names. With six baselines, this yields 100 comparisons per model--baseline pair, or 1,200 pairwise trials over the same 100 prompts. Gemini reports criterion-level preferences for prompt subject and scene, requested action fulfillment, camera-motion compliance, reference-image fidelity, motion integrity, and visual quality and composition. A single overall winner is then chosen from the batch, based on a hierarchical rule: required-motion fulfillment first, then reference fidelity and motion integrity and finally visual quality/composition. As shown in \cref{fig:automated_preference}, our chunk-1 model wins 75\%, 64\%, 62\%, 66\%, 85\%, and 60\% of comparisons against Causal Forcing~\cite{zhu2026causal}, LingBot-World~\cite{team2026advancing}, Causal Forcing++~\cite{zhao2026causal++}, Self-Forcing~\cite{huang2025selfforcing}, Context Forcing~\cite{chen2026contextforcing}, and SANA-WM~\cite{zhu2026sanawm}, respectively. Our chunk-4 model wins 82\%, 76\%, 64\%, 72\%, 88\%, and 74\% against the same baselines.

 \section{Conclusion}

We present Context-Matched Distillation (CMD), a causal DMD framework for few-step autoregressive video generation. CMD addresses the teacher--student information-set mismatch by ensuring that teacher scores cannot depend on future frames or controls. We use the same causal model family for teacher training, student initialization, and inference, avoiding a separate task-specific bidirectional teacher. Within CMD, Prefix Scoring evaluates each target using the student-generated history and controls available when that target was produced. Prefix Corruption makes this supervision more robust when early student rollouts contain severe artifacts. The same causal formulation supports long-video distillation and camera-controlled generation with frame-relative conditioning. Across short and long rollouts, CMD improves generation quality over bidirectional scoring and substantially reduces camera-pose errors. Our ablations isolate the benefits of causal scoring, matched rollout context, and prefix corruption. Overall, CMD shows that matching supervision to the student's causal information state provides a practical foundation for efficient, long-horizon, and controllable video generation.

\newpage
\bibliography{main}
\bibliographystyle{plainnat}

\end{document}